\documentclass[sigconf,nonacm,pbalance]{acmart}

\usepackage{amsmath}
\usepackage{xcolor}
\usepackage{tikz}
\usetikzlibrary{arrows.meta,calc,positioning,fit,backgrounds,shadows}

\definecolor{userfill}{HTML}{FFF4EC}
\definecolor{userstroke}{HTML}{D97706}
\definecolor{sysfill}{HTML}{EEF4FF}
\definecolor{sysstroke}{HTML}{2563EB}
\definecolor{feedfill}{HTML}{F5F3FF}
\definecolor{feedstroke}{HTML}{7C3AED}
\definecolor{phaseone}{HTML}{059669}
\definecolor{phasetwo}{HTML}{0284C7}
\definecolor{phasethree}{HTML}{7C3AED}
\definecolor{phasefour}{HTML}{BE185D}
\definecolor{benefitbg}{HTML}{ECFDF5}
\definecolor{riskbg}{HTML}{FEF2F2}
\definecolor{ecoblue}{HTML}{2563EB}
\definecolor{ecoteal}{HTML}{0D9488}
\definecolor{ecoamber}{HTML}{D97706}
\definecolor{ecoviolet}{HTML}{7C3AED}
\definecolor{ecorose}{HTML}{E11D48}
\definecolor{ecoslate}{HTML}{475569}

\tikzset{
  fig arrow/.style={-{Latex[length=2.4mm,width=2.0mm]}, line width=0.9pt, draw=black!55},
  fig arrow feed/.style={fig arrow, draw=feedstroke, dashed},
  user box/.style={
    rounded corners=5pt, align=center, font=\small\bfseries,
    fill=userfill, draw=userstroke, line width=1pt,
    minimum width=2.35cm, minimum height=0.95cm,
    drop shadow={opacity=0.12, shadow xshift=0.6pt, shadow yshift=-0.8pt}
  },
  sys box/.style={
    rounded corners=5pt, align=center, font=\small\bfseries,
    fill=sysfill, draw=sysstroke, line width=1pt,
    minimum width=2.35cm, minimum height=0.95cm,
    drop shadow={opacity=0.12, shadow xshift=0.6pt, shadow yshift=-0.8pt}
  },
  eco box/.style={
    rounded corners=6pt, align=center, font=\scriptsize\bfseries,
    text=white, inner sep=5pt,
    minimum width=2.05cm, minimum height=0.72cm,
    drop shadow={opacity=0.18, shadow xshift=0.8pt, shadow yshift=-1pt}
  },
  phase card/.style={
    rounded corners=6pt, align=center, font=\scriptsize,
    line width=1pt, inner sep=6pt, text width=2.55cm,
    drop shadow={opacity=0.1, shadow xshift=0.5pt, shadow yshift=-0.7pt}
  },
  tier agent/.style={
    rounded corners=5pt, align=center, font=\scriptsize\bfseries,
    fill=userfill, draw=userstroke, line width=0.9pt,
    text width=2.35cm, minimum height=0.72cm
  },
  tier mem/.style={
    rounded corners=5pt, align=center, font=\scriptsize\bfseries,
    fill=sysfill, draw=sysstroke, line width=0.9pt,
    text width=2.55cm, minimum height=0.72cm
  }
}

\setcopyright{none}
\acmConference[PILA '26 (KDD 2026)]{Personal Intelligence in the Agentic AI Era}{August 9, 2026}{Jeju Island, Republic of Korea}
\acmBooktitle{Personal Intelligence in the Agentic AI Era (PILA '26), co-located with KDD 2026}
\acmDOI{}
\acmISBN{}

\makeatletter
\AtBeginDocument{%
  \fancyhead[LE]{\ACM@linecountL\@headfootfont\footnotesize
    \acmConference@shortname, \acmConference@date, \acmConference@venue}%
  \fancyhead[RO]{\@headfootfont
    \acmConference@shortname,
    \acmConference@date, \acmConference@venue\ACM@linecountR}%
  \fancypagestyle{firstpagestyle}{%
    \fancyhf{}%
    \renewcommand{\headrulewidth}{\z@}%
    \renewcommand{\footrulewidth}{\z@}%
    \fancyhead[L]{\ACM@linecountL\@headfootfont\footnotesize
      \acmConference@shortname, \acmConference@date, \acmConference@venue}%
    \fancyfoot[C]{\if@ACM@printfolios\footnotesize\thepage\fi}%
  }%
}
\makeatother

\title{Toward Personal Intelligence Through Cooperative Observation}

\author{Yashar Talebirad}
\affiliation{%
  \institution{Alberta Machine Intelligence Institute, University of Alberta}
  \city{Edmonton}
  \country{Canada}}
\email{talebira@ualberta.ca}

\author{Osman Jime}
\affiliation{%
  \institution{MacEwan University}
  \city{Edmonton}
  \country{Canada}}
\email{jimeo@mymacewan.ca}

\author{Ali Parsaee}
\affiliation{%
  \institution{University of Alberta}
  \city{Edmonton}
  \country{Canada}}
\email{parsaee@ualberta.ca}

\author{Eden Redman}
\affiliation{%
  \institution{Network for Applied Technology}
  \city{Edmonton}
  \country{Canada}}
\email{eden@nat.ltd}

\author{Yongbin Kim}
\affiliation{%
  \institution{University of Alberta}
  \city{Edmonton}
  \country{Canada}}
\email{yongbin2@ualberta.ca}

\author{Osmar R. Za\"{i}ane}
\affiliation{%
  \institution{Alberta Machine Intelligence Institute, University of Alberta}
  \city{Edmonton}
  \country{Canada}}
\email{zaiane@ualberta.ca}

\begin{document}

\begin{abstract}
A personal AI system needs a model of the user's goals, constraints, and ongoing commitments to plan and act on their behalf, and the quality of that model is bounded by what the system can observe. Broader observation does not by itself improve assistance because a bounded system must select and compress information for the task at hand. We argue that this observation bottleneck has a cooperative structure: the system builds a partial model of the user's changing life, the user evaluates its actions, and the user's consent and control shape what it can observe next. Useful and inspectable behavior can give users a reason to maintain or expand the observation channel, while failures can lead them to correct, narrow, revoke, or abandon it. We use the term \emph{cooperative observation} for this feedback loop among usefulness, trust, and future access, and propose it as a framework for personal intelligence. We report a preliminary single-subject account from Organizm, a prototype used over six months, and outline evaluation directions for measuring how observation quality shapes personal AI.
\end{abstract}


\maketitle
\raggedbottom

\section{Introduction}

Personal AI is moving from a speculative idea to an active engineering target. Open-source projects such as OpenClaw and NanoClaw\footnote{OpenClaw: \url{https://github.com/openclaw/openclaw}. NanoClaw: \url{https://github.com/nanocoai/nanoclaw}.} aim to give users personal assistants that run across devices and tools, often emphasizing local control and inspectability. Commercial systems are also converging on persistent user memory, voice interfaces, and agentic tool use. At the same time, consumer hardware is moving closer to continuous capture through smart glasses, wearable audio, health sensors, and eventually neural interfaces.

The need for user context links these developments. Personal intelligence depends on knowledge of the user's goals, constraints, habits, and ongoing commitments. Relevant context lets a model plan, catch inconsistencies, and support decisions across time. The system cannot directly access the user's life, so its personal model is bounded by a lossy stream of reports, sensor readings, and interactions. Agent software and sensing hardware are approaching this observation bottleneck from opposite sides: agents need richer context to act personally, while new devices capture context that becomes useful once an agent can act on it.

Much of the relevant context is already being collected: smart watches and health platforms record sleep, activity, and stress, signals directly useful for personal assistance, but access is often restricted by platform-specific exports, APIs, and permission systems. An assistant chosen by the user therefore lacks access to signals that would help it most, and the user may have no practical way to transfer data describing their own body and days. Richer sensing devices expand what can enter the observation stream, but leave open how that stream's configuration is governed and which objective is optimized over the resulting history. Local processing can support cooperative observation by keeping channel configuration and memory under the user's control, allowing continuous capture without continuous external disclosure.

Computational systems already model people at scale. Recommender systems, social feeds, and advertising infrastructures infer preferences and vulnerabilities from behavioral traces, and the objectives they serve, such as engagement, retention, and revenue, belong to the platform rather than to the person being modeled \citep{stray2020aligning,stray2021optimizingforaligningrecommender,stray2024building,boerman2017online}. Richer observation flowing into systems of this kind would deepen surveillance, manipulation, and dependency. The same modeling capacity, aimed at goals the user sets and evaluated under the user's own reward signal, could instead help a person understand and steer their own life amid competing models built for other objectives, if the observation channel and the objective remain under that person's control. Personal intelligence depends on structural conditions that make observation cooperative.

In a partially observable decision problem, an assistant needs three components: a model of the relevant state, a signal that defines whether its actions are good, and an observation process that determines what it can learn next. In personal AI, all three are anchored in the same person: the state concerns the user's goals, constraints, commitments, and other parts of life relevant to the task; the user's own evaluation supplies feedback; and their willingness to share or withhold data governs much of the observation process. Most personalization systems split these roles across parties: platforms often define the objective and govern how behavioral data is used. In user-owned personal AI, however, the person being modeled can define success and control future access to their data.

We propose \emph{cooperative observation} as a framework for user-governed personal intelligence. Evaluation therefore considers both the quality of assistance and changes to the observation channel. We use an observation-channel spectrum to illustrate how available signals may broaden from deliberate reports to continuous and neural observation. We also frame \emph{task-conditioned context allocation} as the problem of selecting and representing user information for a decision under processing and disclosure constraints. We then describe Organizm, a prototype built on user-owned files, explicit memory, and feedback-driven planning.

\section{The Rise of Personal AI Systems}

Large language models can act as general-purpose interfaces over tools, files, and communication channels. Reinforcement learning from human feedback also shows that model behavior can change substantially when training is shaped by human preferences and instructions \citep{christiano2017deep,ouyang2022training}. These capabilities support assistants that plan, maintain memory, incorporate feedback, and use tools across sustained interactions.

Recent agent systems combine language models with memory, planning, and reflection, showing how persistent context can change the behavior of language-model systems \citep{park2023generative}. Projects such as OpenClaw and NanoClaw extend this in a user-facing direction, offering cross-platform personal agents with persistent memory, inspectable execution, and local or user-chosen models. OpenClaw-style ecosystems already share reusable skills that extend what an agent can do.\footnote{OpenClaw skills documentation: \url{https://docs.openclaw.ai/tools/skills} (accessed July 2026).} Observation and memory can be modular as well, with each module exposing what it collects and retains. The context a personal assistant needs most, covering goals, habits, health, and finances, is also the context users are least willing to place in services they cannot inspect.

Model capability is also becoming more deployment-efficient. An analysis of 51 open-source base models found that maximum capability per parameter increased sharply over time, while architecture-aware scaling experiments show that models can gain accuracy and inference throughput under matched training budgets \citep{xiao2025densing,bian2026scaling}. As this frontier advances, personal agents can interpret and act on a fixed history more effectively while keeping more inference on user-controlled devices. This complements observation-channel growth, which can supply additional user-specific evidence.

A multi-week field deployment of a context-aware LLM chatbot found that participants who communicated more frequently about their context received more specific and relevant suggestions. Inaccurate recommendations discouraged some users from further interaction, while privacy concerns and reluctance to criticize the chatbot also limited context sharing and correction \citep{xu2025goals}. The support that the users wanted changed during the deployment, moving from action discovery toward planning, tracking, reflection, motivation, and accountability.

Devices are also becoming better at capturing the user's world. Smart glasses, wearable audio, and health sensors already expose richer context than typed text, and neural interfaces have begun to decode attempted speech directly from neural activity \citep{willett2023highperformance}. These devices widen the observation channel available to personal AI.

Personal intelligence also requires continual learning. The user's state is non-stationary as goals change, preferences drift, and constraints appear. Continual-learning research studies how a system can absorb new information without overwriting what it already knows \citep{parisi2019continual}, and a personal AI faces that trade-off in its user model. Persistent memory and feedback incorporation help distinguish stable preferences from temporary noise.

The same dependence on context also operates within a single user-system relationship. A user who begins by sharing only explicit tasks may later share priorities, constraints, or project history after receiving useful plans. The user may connect calendars, wearables, or continuous capture as trust grows, then narrow or close the channel after failures. The framework treats user control over these changes as a requirement.

\section{The Cooperative Observation Framework}

A personal AI acts under partial observability \citep{kaelbling1998planning} and under uncertainty about what its user values \citep{hadfieldmenell2016cirl}. It must infer the task-relevant parts of the user's changing life while learning their goals and preferences from feedback.

Standard partially observable models specify the observation process in advance, while systems with controllable sensing may allow the agent to choose what to observe. In user-owned personal AI, much of this choice belongs to the user, who can enable, withhold, or revoke data sources in response to system performance. The same person is therefore the subject of the state model, the evaluator of actions, and the controller of the observation channel. In the formal setup below, uppercase letters denote random variables and lowercase letters denote realized values.

The framework has three parts. A one-step decision model specifies the user's state, the system's observations and actions, and the user's immediate and reflective evaluations. An information-theoretic bound then describes the ceiling imposed by the system's observation history. Finally, a feedback loop lets the user's experience of the action change what the system may observe next.

\subsection{A One-Step Decision Model}

\paragraph{State.}
We use $S_t \in \mathcal{S}$ for the user's evolving state: goals, constraints, commitments, ongoing priorities, external facts such as deadlines and meetings, and internal variables such as motivation, overload, and preference drift. A particular task or outcome determines which components of this state are relevant to choosing or evaluating an action: a calendar may reveal a scheduling conflict while carrying little information about motivation. This creates an information-bottleneck problem: a bounded system must preserve the information needed for the decision while compressing a broader observation history \citep{tishby1999information}. Section~\ref{sec:task-conditioned-compression} develops this connection. Internal variables are latent and must be inferred from reports or behavior, whereas external facts become available through connected data sources or user reports.

\paragraph{Observation.}
At each step, the system receives an observation $O_t$ about $S_t$. The channel configuration $\Phi_t$ specifies which data sources are active and how much the user engages with each, and it can change over time under the user's control. For simplicity, we omit additional dependencies on prior actions and interaction history.

\paragraph{Action.}
Let $H_t = (O_{1:t}, \Phi_{1:t})$ denote the system's information history, comprising the observations it has received and the channel configurations through which they arrived. Requests and queries are themselves observations, so $H_t$ carries what the system has been asked as well as what it has learned, and derived memory such as summaries and indexes is a function of the same history. The system selects actions according to a policy $\pi$, with $A_t = \pi(H_t)$. The policy may encode arbitrary general knowledge, but its user-specific information can come only from $H_t$.

\paragraph{Reward.}
The user evaluates the action through ratings, acceptance decisions, or corrections. We write $R_t \in \mathbb{R}$ for a scalar summary of the immediate evaluation available to the system. Corrections can also supply information for subsequent decisions. In this framework, $R_t$ represents user-provided evaluation rather than platform-specific objectives such as engagement, retention, or click-through rate. Preference-learning systems show one route for incorporating such feedback into model behavior \citep{christiano2017deep,ouyang2022training}.

\paragraph{Approval and reflective benefit.}
Immediate approval can differ from reflective assessment. We write $U_t \in \mathbb{R}$ for a scalar summary of the user's reflective assessment of whether the action was beneficial, evaluated at a specified horizon. Unlike $R_t$, $U_t$ is not assumed to be available to the system during interaction because relevant consequences may become apparent only later. When $R_t$ is optimized as a proxy for $U_t$, this becomes an instance of Goodhart's law: $R_t$ can rise without a corresponding gain in reflective benefit \citep{goodhart1984monetary}. Section~\ref{sec:evaluation-directions} outlines possible operational proxies for these quantities.

\subsection{The Observation Ceiling}

For fixed model parameters and policy, the system acts on its information history, and the information flow forms a Markov chain $S_t \to H_t \to A_t$. By the data processing inequality \citep{cover2006elements}:
\begin{equation}
I(S_t;\, A_t) \leq I(S_t;\, H_t).
\end{equation}

The left side measures how much the system's action varies with the user's true state under the chosen state distribution. An action independent of the state carries zero mutual information, while greater state dependence can produce higher mutual information. Mutual information alone does not imply utility, since an action can encode state without helping the user. The right side measures how much the information history reveals about the current state. The inequality says that state-dependent personalization cannot exceed the information the history carries about the user. The inequality gives only an upper bound: showing that observation limits a particular system requires comparing the same system under different observation conditions.

The bound separates two sources of improvement. At a given step, changing the model or policy can help the system use information already present in $H_t$, but it cannot add user-specific information to that history. Expanding the observation channel can raise $I(S_t; H_t)$ and therefore the upper bound on state-dependent personalization. Which parts of the history are informative also depends on the task and time: old observations may lose relevance as the user's state shifts, and different channels reveal different parts of the user's life. The observation-channel ablation in Section~\ref{sec:evaluation-directions} therefore holds the model fixed while varying the channel.

\subsection{The Cooperative Feedback Loop}

The user's experience can then change future observability. After seeing the system's action, the user may maintain, expand, narrow, or revoke the channel configuration for the next step, $\Phi_{t+1}$. This decision depends on immediate evaluation, expectations about longer-term benefit, trust, effort, and privacy cost, and we leave the update rule unspecified. These factors can conflict: a user may find a system useful and still refuse to share health data when the privacy cost outweighs the marginal benefit.

Channel configuration affects what the system can learn and how well it can help. Performance then affects the user's willingness to maintain, expand, narrow, or revoke the channel. The user's benefit also depends on who sets the objective and governs the channel. For instance, a richer user model can help a student notice weeks of drift from stated goals. Under a platform objective, the same model can target advertising at that drift and deepen dependence.

At the level of individual updates, we define a channel update as \emph{cooperative} when it remains under user control and improves reflective benefit. In contrast, expansion is \emph{extractive} when access or engagement increases without such improvement. Continued use or disclosure alone does not establish trust or reflective benefit.

\begin{figure}[t]
  \centering
  \begin{tikzpicture}[
      label/.style={font=\scriptsize, fill=white, inner sep=1.5pt, rounded corners=1pt},
      every node/.style={transform shape}
    ]
    \def\R{1.95}
    \def\Rf{2.75}
    \node[user box, minimum width=2.2cm] (obs) at (90:\R) {User\\observation};
    \node[sys box, minimum width=2.2cm] (model) at (0:\R) {System\\memory/model};
    \node[sys box, minimum width=2.2cm] (action) at (270:\R) {System\\action};
    \node[user box, minimum width=2.2cm] (eval) at (180:\R) {User\\evaluation};

    \draw[fig arrow] (obs.east) to[bend left=8]
      node[label, above right=-2pt, pos=0.5] {$O_t$} (model.north);
    \draw[fig arrow] (model.south) to[bend left=8]
      node[label, below right=-2pt, pos=0.5] {$H_t$} (action.east);
    \draw[fig arrow] (action.west) to[bend left=8]
      node[label, below left=-2pt, pos=0.5] {$A_t$} (eval.south);
    \draw[fig arrow] (eval.north) to[bend left=8]
      node[label, above left=-2pt, pos=0.5] {$R_t$} (obs.west);

    \draw[fig arrow feed]
      (235:\Rf) arc (235:305:\Rf);
    \node[font=\scriptsize, text=feedstroke, align=center] at (270:\Rf+0.4)
      {Value, trust, effort, privacy $\rightarrow \Phi_{t+1}$};
  \end{tikzpicture}
  \caption{Cooperative observation as a recurring loop. User evaluation affects the next observation channel, so observability is partly earned through usefulness, trust, and control.}
  \Description{A circular four-node diagram with warm user nodes and cool system nodes, labeled with O_t, H_t, A_t, and R_t, plus a dashed purple feedback arc for channel updates.}
  \label{fig:cooperative-loop}
\end{figure}
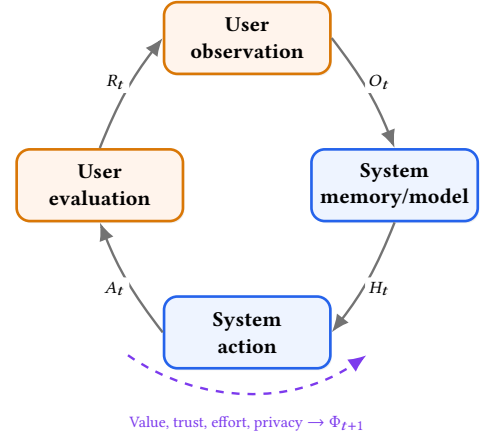

We define \emph{cooperative observation} as a feedback process in which a personal AI system and its user jointly adjust the system's access to task-relevant user context because doing so benefits the user. Figure~\ref{fig:cooperative-loop} summarizes the process: the user provides a task, correction, or sensor stream; the system updates its working model and acts; and the user evaluates the action before sharing more, correcting the model, or withholding data.

Cooperative observation has a cold-start problem: personal context can make the system more useful, but users may wait for evidence of usefulness and trustworthiness before sharing it. In a multi-week deployment of a context-aware chatbot, participants who communicated more context received more specific and relevant suggestions, while inaccurate suggestions and privacy concerns discouraged further interaction or disclosure \citep{xu2025goals}. The GOD model addresses the same problem by rewarding users with platform tokens for data that improves their assistant \citep{pinai2025god}. Cooperative observation changes this incentive loop by keeping evaluation and access decisions with the person being modeled. Useful assistance may itself motivate the user to maintain or expand the channel when the expected value of an added source for a particular task outweighs its effort and privacy costs. Viewed as a dynamical system, the loop may settle into a stable set of channels or remain history-dependent, for example when repeated success is needed to earn access but one serious failure is enough to lose it. Characterizing these regimes, including revocation and abandonment, is an open empirical question.

\section{Observation Channels and the Complementary Self-Model}
\label{sec:observation-channels}

Observation fidelity concerns how broadly, continuously, and directly information about the user's evolving state reaches the system. It can increase through more frequent observations, additional modalities, or more direct measurements. A continuous audio-visual stream may therefore raise the information ceiling beyond sporadic reports. Yet additional bandwidth need not carry information relevant to a particular task, so observation fidelity alone does not imply usefulness. Section~\ref{sec:task-conditioned-compression} explains how a bounded system selects and compresses observations for a particular decision.

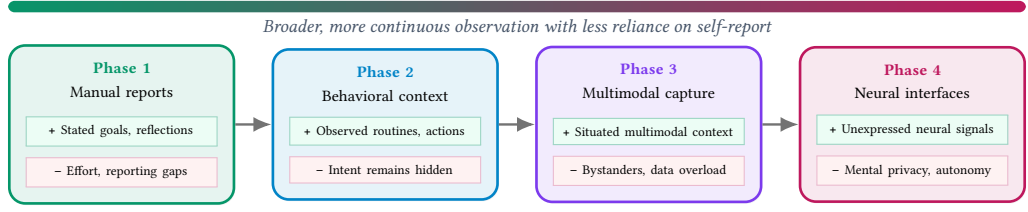
\begin{figure*}[t]
  \centering
  \begin{tikzpicture}
    \node[phase card, fill=phaseone!10, draw=phaseone] (p1) {
      {\color{phaseone}\textbf{Phase 1}}\\[2pt]
      Manual reports\\[5pt]
      \fcolorbox{phaseone!40}{benefitbg}{\parbox{2.30cm}{\centering\tiny\textbf{+} Stated goals, reflections}}\\[3pt]
      \fcolorbox{phasefour!30}{riskbg}{\parbox{2.30cm}{\centering\tiny\textbf{--} Effort, reporting gaps}}
    };
    \node[phase card, fill=phasetwo!10, draw=phasetwo, right=0.48cm of p1] (p2) {
      {\color{phasetwo}\textbf{Phase 2}}\\[2pt]
      Behavioral context\\[5pt]
      \fcolorbox{phaseone!40}{benefitbg}{\parbox{2.30cm}{\centering\tiny\textbf{+} Observed routines, actions}}\\[3pt]
      \fcolorbox{phasefour!30}{riskbg}{\parbox{2.30cm}{\centering\tiny\textbf{--} Intent remains hidden}}
    };
    \node[phase card, fill=phasethree!10, draw=phasethree, right=0.48cm of p2] (p3) {
      {\color{phasethree}\textbf{Phase 3}}\\[2pt]
      Multimodal capture\\[5pt]
      \fcolorbox{phaseone!40}{benefitbg}{\parbox{2.30cm}{\centering\tiny\textbf{+} Situated multimodal context}}\\[3pt]
      \fcolorbox{phasefour!30}{riskbg}{\parbox{2.30cm}{\centering\tiny\textbf{--} Bystanders, data overload}}
    };
    \node[phase card, fill=phasefour!10, draw=phasefour, right=0.48cm of p3] (p4) {
      {\color{phasefour}\textbf{Phase 4}}\\[2pt]
      Neural interfaces\\[5pt]
      \fcolorbox{phaseone!40}{benefitbg}{\parbox{2.30cm}{\centering\tiny\textbf{+} Unexpressed neural signals}}\\[3pt]
      \fcolorbox{phasefour!30}{riskbg}{\parbox{2.30cm}{\centering\tiny\textbf{--} Mental privacy, autonomy}}
    };

    \draw[fig arrow] (p1.east) -- (p2.west);
    \draw[fig arrow] (p2.east) -- (p3.west);
    \draw[fig arrow] (p3.east) -- (p4.west);

    \shade[left color=phaseone, right color=phasefour, rounded corners=2.5pt]
      ([yshift=12pt]p1.north west) rectangle ([yshift=17pt]p4.north east);
    \node[font=\footnotesize\itshape, text=ecoslate, above=1.05cm of $(p2)!0.5!(p3)$]
      {Broader, more continuous observation with less reliance on self-report};
  \end{tikzpicture}
  \caption{Observation-channel spectrum for personal AI. Later phases can broaden the observation stream through more continuous modalities that rely less on self-report, while also increasing privacy, consent, and autonomy risks.}
  \Description{A four-card spectrum in green, blue, purple, and rose, each with an observation benefit and a central limitation or risk, connected left to right beneath a gradient bar indicating broader, more continuous observation with less reliance on self-report.}
  \label{fig:fidelity-spectrum}
\end{figure*}

We use four phases to mark broad positions on this spectrum, from deliberate self-report to continuous and neural observation (Figure~\ref{fig:fidelity-spectrum}). The phases are not mutually exclusive categories: a system may combine channels from several phases. Moving toward later phases can increase the information available to the system, while also raising privacy, consent, and governance demands.

\paragraph{Phase 1: Manual reports.}
At the most deliberate end of the spectrum, the user writes or says what matters: tasks, goals, and reflections. This channel directly captures stated intentions, but it is low bandwidth, effortful, and prone to reporting gaps. With consistent reporting, it supports planning, goal tracking, and memory. Personal informatics systems provide a familiar example \citep{li2010stage}.

\paragraph{Phase 2: Automated behavioral context.}
The system observes calendars, messages, health data, or screen activity. This phase reduces reporting burden and provides evidence about what the user actually did, supporting habit detection and behavior-intention gap analysis. It also increases privacy risk because passive traces reveal more than the user may consciously intend to share.

\paragraph{Phase 3: Continuous multimodal capture.}
Smart glasses, always-on audio, cameras, and richer wearables capture situations the user may not explicitly report. Earlier lifelogging and wearable-camera systems, including MyLifeBits and SenseCam, explored related ideas of personal capture and retrospective memory support \citep{gemmell2006mylifebits,hodges2006sensecam}. This phase can support context-aware assistance and fine-grained situation recognition. However, continuous capture also collects information about people other than the user, making bystander privacy, consent, and transparency about inferred information central design requirements.

\paragraph{Phase 4: Neural or near-neural interfaces.}
Neural interfaces add an inward-facing source to the observation stream by measuring neural activity directly. Such signals may eventually provide information about internal processes, including inner speech, perception, attention, or affect, that external observation cannot directly capture. Existing demonstrations are much narrower: they decode attempted speech in a participant with paralysis \citep{willett2023highperformance}, a small vocabulary of internal speech \citep{wandelt2024internal}, and semantic content from perceived and imagined speech \citep{tang2023semantic}. Neural signals could complement multimodal capture by linking internal processes to the external situations in which they occur. However, current systems provide neither unrestricted nor transparent access to thought, and open-ended access to thought remains the motivating limit case. Because such a channel could extend beyond deliberate disclosure, it heightens concerns about mental privacy, autonomy, identity, and control.

\subsection{The Complementary Self-Model}

Self-tracking turns observations about daily life into persistent records that users can revisit. Personal informatics systems and the Quantified Self movement use such records to support self-knowledge \citep{li2010stage}. Across 138 randomized studies, interventions designed to increase monitoring of goal progress improved goal attainment, with larger effects when information obtained through monitoring was recorded or reported \citep{harkin2016monitoring}. In a personal health informatics study, participants used generative AI to analyze diverse tracking data, explore relationships between health measures and daily behavior, validate existing practices, and identify new health goals \citep{chopra2025engagements}. A personal AI can extend these practices by relating such signals across time and bringing them into planning and action, for example by weighing a poor night's sleep against deadlines and project history before proposing a schedule.

Over time, such a system can build a model that complements the user's existing self-knowledge. Users know some aspects of their present experience firsthand, whereas an external system must infer those aspects from reports and observations. Repeated records can reveal changes over time that retrospective recall may miss \citep{bolger2003diary}. For personal AI, these patterns may include recurring problems, shifts in time allocation, and gradual drift between schedules and stated goals. Some relevant knowledge is also difficult to put into words \citep{polanyi1966tacit}, limiting what manual reports can capture.

Persistent memory can support this longer view by comparing reports and records across weeks or months. It can keep deadlines, recurring bottlenecks, and neglected projects in view, and can surface a sustained gap between stated priorities and recorded time allocation. Broader or more continuous channels can strengthen these comparisons: calendars and activity traces add evidence about scheduled and observed behavior, while multimodal capture preserves context absent from written reports. A sufficiently calibrated self-model could support counterfactual planning by estimating how candidate plans might interact with the user's likely behavior, but blindly treating predicted behavior as an endorsed objective could reinforce habits the user wants to change, steer their choices, or justify actions without authorization.

A complementary self-model can become a tool the user relies on for remembering, reflecting, and deciding, allowing it to function as part of an extended cognitive process \citep{clark1998extended}. The user must retain control over what the system observes, remembers, and does. The Good Regulator theorem provides a second motivation: effective regulation depends on a model of what is being regulated \citep{conant1970good}. For personal assistance, this points toward modeling the goals, commitments, and constraints relevant to the user's decisions.

\subsection{Task-Conditioned Selection and Compression}
\label{sec:task-conditioned-compression}

Broader observation can leave the system with more history than it can use in any single decision. The big world hypothesis treats this mismatch as fundamental: an agent cannot fully perceive a world much larger than itself or represent the correct response to every state, so it must rely on approximation \citep{javed2024bigworld}. A personal AI must therefore select and compress its observation history for the task at hand. The information bottleneck formalizes the resulting trade-off: a compact representation should preserve information about the decision-relevant target while retaining as little of the full input as possible \citep{tishby1999information}. The relevant target changes across users, tasks, and time, and several targets may matter at once. Compression for one target can therefore remove evidence needed for another, so systems need task-specific summaries and, where the user's policy allows it, a path back to the source material. Existing work asks whether retrieved context is sufficient to answer a query \citep{joren2025sufficient}. Personal AI poses a broader \emph{task-conditioned context allocation} problem: deciding how much user information to make available, from which sources, and in what representation and abstraction so that it supports the task within processing and disclosure constraints.

Systems often seek broader coverage by combining channels with different formats and timescales. This can provide evidence for decisions that span several parts of the user's life, but the system must decide which sources to consult, at what resolution, and how to reconcile observations collected on different timescales. An information-theoretic account of bounded rationality shows why hierarchy can help: trading expected utility against information-processing cost can induce abstractions at different levels of a decision hierarchy \citep{genewein2015bounded}. The \emph{Society of Mind} provides a complementary architectural view in which specialized processes interact to produce intelligence \citep{minsky1986society}. These perspectives suggest a recurring response to boundedness across the user-system boundary: The user externalizes experience into persistent records, and bounded agents select, compress, and route those records at task granularity.

\section{Organizm: A Prototype for Cooperative Observation}
\label{sec:organizm}

Organizm is a prototype for cooperative observation that has been used in practice over multiple months. It currently runs on manual reports, user-owned files, and lightweight calendar integration. Section~\ref{sec:deployment-sketch} describes a six-month single-subject deployment, including changes in reporting frequency and planning inputs together with examples of user correction. Architecturally, Organizm implements the \emph{society of mind} view \citep{minsky1986society} through a hierarchical multi-agent system coupled to a user-owned file hierarchy.

Organizm uses hierarchical externalization and task-conditioned selection across the user-system boundary. The user externalizes personal context into persistent records organized at multiple granularities. Because no single agent can use the full record at once, the system selects task-appropriate working sets and assigns specialized agents to corresponding scopes. The coupled hierarchy therefore extends the user's capacity while managing the system's own capacity limits. It also combines file-native memory, coupled agent and information hierarchies, and a feedback loop that updates plans and governs progressive integration.

\subsection{File-Native Memory and Index-First Traversal}

The system stores daily logs, project notes, and session summaries as ordinary files in a user-owned file hierarchy. This local-first design makes user control, data ownership, and software continuity architectural requirements while allowing the user to inspect, edit, or delete the memory directly \citep{kleppmann2019localfirst}.

Folder-level and project-level index files summarize what lives below them, allowing an agent to route through compressed representatives before reading full files \citep{talebirad2026hierarchical}. An agent starts with a coarse index and follows it to more detailed files only when the task requires them.

\subsection{Coupled Agent and Information Hierarchies}

Within the system, different tasks require different context scopes. Immediate note processing does not need the full life history, while long-term reflection needs goals, values, and historical summaries.

\begin{figure}[t]
  \centering
  \resizebox{\columnwidth}{!}{%
  \begin{tikzpicture}[
      col label/.style={font=\footnotesize\bfseries, text=ecoslate},
      link/.style={fig arrow, draw=userstroke!70},
      tier agent/.style={
        rounded corners=5pt, align=center, font=\scriptsize\bfseries,
        fill=userfill, draw=userstroke, line width=0.9pt,
        text width=2.4cm, minimum height=0.68cm
      },
      tier mem/.style={
        rounded corners=5pt, align=center, font=\scriptsize\bfseries,
        fill=sysfill, draw=sysstroke, line width=0.9pt,
        text width=2.55cm, minimum height=0.68cm
      }
    ]
    \node[tier agent] (a4) {Long-term\\architecture};
    \node[tier agent, below=0.28cm of a4] (a3) {Weekly\\strategy};
    \node[tier agent, below=0.28cm of a3] (a2) {Daily\\coaching};
    \node[tier agent, below=0.28cm of a2] (a1) {Immediate\\note processing};

    \node[tier mem, right=0.7cm of a4] (m4) {Goals, values,\\slow summaries};
    \node[tier mem, below=0.28cm of m4] (m3) {Project summaries,\\weekly reviews};
    \node[tier mem, below=0.28cm of m3] (m2) {Daily logs,\\deadlines, inbox};
    \node[tier mem, below=0.28cm of m2] (m1) {Session traces,\\new notes};

    \node[col label, above=0.35cm of a4] {Agent personas};
    \node[col label, above=0.35cm of m4] {Information tiers};

    \foreach \x/\y in {a1/m1,a2/m2,a3/m3,a4/m4} {
      \draw[link] (\x.east) -- (\y.west);
    }

    \begin{scope}[on background layer]
      \node[fit=(a1)(a4), inner sep=6pt, rounded corners=8pt,
        fill=userfill, fill opacity=0.45, draw=userstroke, line width=0.9pt, dashed, draw opacity=0.7] {};
      \node[fit=(m1)(m4), inner sep=6pt, rounded corners=8pt,
        fill=sysfill, fill opacity=0.45, draw=sysstroke, line width=0.9pt, dashed, draw opacity=0.7] {};
    \end{scope}

    \draw[fig arrow, draw=ecoamber]
      ([xshift=-0.55cm]a4.west) -- node[left, font=\scriptsize, text=ecoamber, pos=0.5]
        {Priorities $\downarrow$} ([xshift=-0.55cm]a1.west);
    \draw[fig arrow feed]
      ([xshift=0.55cm]m1.east) -- node[right, font=\scriptsize, text=feedstroke, pos=0.5]
        {Summaries $\uparrow$} ([xshift=0.55cm]m4.east);
  \end{tikzpicture}%
  }
  \caption{The prototype couples agent personas to matching information tiers. Higher levels preserve abstraction, while lower levels keep current observations close to action.}
  \Description{Two color-framed vertical columns of four tiers each, linked horizontally, with upward summary flow and downward priority flow between the columns.}
  \label{fig:coupled-hierarchy}
\end{figure}
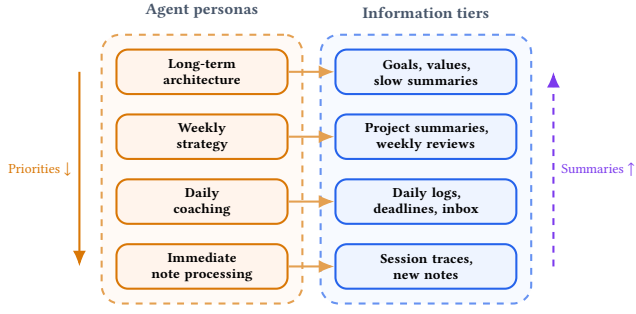

The system is based on a hierarchical multi-agent interface that is coupled to a hierarchical information store \citep{talebirad2026hierarchical}. Specialized personas operate at scopes ranging from immediate processing to long-term architecture. The agent hierarchy and the memory hierarchy are aligned by timescale and abstraction level (Figure~\ref{fig:coupled-hierarchy}).

This coupling implements task-conditioned context allocation from Section~\ref{sec:task-conditioned-compression}: each agent initially loads the information tier matched to its task, while index-first traversal supplies finer detail when that tier is insufficient. Longer-horizon goals, values, and commitments guide project plans and daily tasks, while evidence from lower-level observations and outcomes flows upward through progressively coarser summaries to support reflection and revision. This aligns goal horizon with agent scope and memory granularity. The same design choice has been tested in an ML-engineering system, where tiered loading outperformed flat loading while using fewer output tokens \citep{kim2026haste}.

\subsection{Feedback, Integration, and Control}
\label{sec:feedback-integration}

The prototype maintains daily plans, weekly reviews, and user corrections to improve future planning. Daily and weekly files compare planned activity against actual activity, turning self-report into a recurring observation channel.

Cooperative observation here depends on what the user reports after acting. Useful plans give the user a reason to report outcomes, whereas generic plans can make reporting feel like overhead and weaken the loop. Corrections provide calibration: rejecting overambitious daily plans teaches the system how to plan, while feedback on anxiety-inducing reminders teaches deadline escalation.

The same feedback loop can guide progressive integration across the observation-channel spectrum in Section~\ref{sec:observation-channels}. Organizm's current integrations remain within Phases~1--2, as described in Section~\ref{sec:deployment-sketch}. Future additions (e.g., health summaries or wearable signals) should have a user-endorsed purpose and a visible control boundary.

Later phases would also increase data volume and sensitivity, affecting where inference should run. High-volume streams may make remote inference expensive, while sensitive streams may favor local processing. A later version could assign each channel to a specialized agent that compresses its stream before broader planning and escalates when confidence is low, when a task crosses domains, or when the user requests broader review. The user would retain control over which channels are enabled.

\subsection{Six-Month Deployment Sketch}
\label{sec:deployment-sketch}

We report an illustrative sketch of one author's first six months using Organizm (January--June 2026).\footnote{The subject is an author and consented to publication of these aggregate records.} Manual logging varied early and became near daily by May--June (Table~\ref{tab:deployment-logs}), while weekly reviews and monthly summaries were added over the same period. These changes are consistent with cooperative observation, but the retrospective single-subject record cannot identify their causes or support generalization; Section~\ref{sec:evaluation-directions} describes the multi-subject studies needed to test the proposed feedback loop.

\begin{table}[t]
  \centering
  \caption{Daily log files per month in the single-subject Organizm deployment (2026).}
  \label{tab:deployment-logs}
  \begin{tabular}{lcccccc}
    \hline
    Month & Jan & Feb & Mar & Apr & May & Jun \\
    \hline
    Daily logs & 24 & 9 & 12 & 21 & 31 & 30 \\
    \hline
  \end{tabular}
\end{table}

Over the same period, the observation channel expanded within Phases~1--2 of the observation-channel spectrum without wearables or continuous capture. Calendar events entered the daily planning loop first, followed by a shared deadline registry, weekly and monthly review layers, and a finance channel loaded by a specialized agent. Each addition was voluntary and tied to a specific purpose, with the resulting information kept in user-owned files that could be inspected or deleted. Corrections also changed the stored model and later plans: in one episode, the user corrected a cost model that omitted a recurring expense; in another, the system revised a daily plan after checking its interpretation of the user's commitments against the live calendar. These corrections updated memory and future planning via the feedback loop in Section~\ref{sec:feedback-integration}.

\section{Evaluation Directions}
\label{sec:evaluation-directions}

We propose three complementary studies to evaluate cooperative observation. An observation-channel ablation can test whether task-relevant context improves assistance with a fixed model. Additionally, a longitudinal deployment can test whether assistance quality affects what users later share, correct, or revoke. Finally, a self-model study can examine changes in the system's knowledge and the user's self-knowledge.

\paragraph{Observation-channel ablations.}
To test whether observation improves assistance, an ablation study would hold the model, tools, prompt template, and sampling settings fixed. Before evaluation, researchers would assemble a suite of real, participant-specific tasks with checkable constraints or outcomes, grouped into predefined categories such as scheduling, prioritization, planning, and forecasting. Each task would be evaluated under five separate context conditions: no personal context, a static profile, recent manual logs, behavioral or calendar traces, and a combined condition containing the profile, recent logs, and available traces.
Prompt length would be matched where possible, with irrelevant context of similar length included as a control for the effect of additional tokens. Responses would be scored using criteria defined in advance for each task.
When human judgment is needed, raters would see only the task and response. With graded channel configurations, the same design could test whether task performance changes smoothly, saturates, or exhibits task-specific thresholds as observation broadens.

For each scored claim in a response, the analysis would record whether its supporting information came from the task request, the context supplied in that condition, or an inference from other evidence. A claim would count as an inference only when the target fact is absent from the task and supplied context and can be checked against an independent record. The no-context condition measures performance from the task request alone. A provenance check records which supplied sources support the response and flags cases in which the target already appears in the input. Across conditions, the results would show which context sources improve which kinds of tasks.

\paragraph{Usefulness and channel change.}
To test whether assistance affects future observation, a longitudinal deployment would record perceived usefulness and task performance, then track what the user shares, corrects, or revokes over subsequent interactions. For each source, the study would record whether the user adds it, keeps it unchanged, reduces the detail it provides, or removes it, since users may weigh benefit, reporting effort, and privacy cost differently across sources. Reporting these changes for each source preserves channel-specific patterns within the overall sharing trend.

The study should also track changes in the types of tasks users bring to the system, since the feedback loop may change the task distribution the system observes.

\paragraph{Complementary self-model evaluation.}
System knowledge can be evaluated through paired predictions of later, verifiable outcomes, with one prediction based on the task alone and another using the task plus prior personal records. User self-knowledge can be evaluated through changes in the accuracy of the user's forecasts of time allocation and project delays, along with assessments of consistency between planned work and stated goals.

\section{Safety and Governance}
\label{sec:safety-governance}

Richer observation increases both what a personal AI can do and the consequences of misuse or failure. Safety depends on which inferences are retained, whether observation channels can be manipulated, how feedback shapes behavior, how agents share authority, and whether users can withdraw access and retain their data. Because the person being modeled also supplies feedback and controls access, personal AI may provide a small-scale testbed for alignment under uncertainty about user values \citep{hadfieldmenell2016cirl}. The framework distinguishes immediate approval, reflective benefit, and decisions about continued access.

\paragraph{Observation privacy and integrity.}
Raw records can support sensitive conclusions. For instance, a user may knowingly share sleep data without anticipating that the system could infer a health condition. A breach can therefore expose both the source data and the conclusions drawn from it, and retained inferences should receive the same privacy protection as their sources. Continuous audio and visual systems also capture bystanders, whose consent cannot be supplied by the primary user.

Neural and other high-bandwidth channels also separate permission to collect a signal from permission to derive a particular inference. Access to raw neural activity should not be treated as blanket authorization to decode inner speech, attention, affect, or other latent attributes. Controls should therefore be scoped to particular purposes and types of inference, disclose what was inferred and retained, and support pausing, deletion, and revocation for both raw signals and derived representations.

Furthermore, every added channel broadens the attack surface. Red-teaming work on autonomous agents shows that malicious content in the observation stream can redirect behavior even when the user's instruction is sound \citep{zhao2026clawtrap}.\footnote{Reported failures include unauthorized bulk email deletion after an agent lost the user's approval constraint during context compaction; see Kevin Okemwa, ``Meta's safety director handed OpenClaw AI agents the keys to her emails,'' \url{https://www.windowscentral.com/artificial-intelligence/meta-summer-yue-director-openclaw-ai-email-deletion} (accessed July 2026).} Shared skills create another path because their instructions enter the agent's context and may request access to tools or data.\footnote{Cisco researchers demonstrated a highly ranked malicious skill that combined direct prompt injection with silent data exfiltration; see Amy Chang and Vineeth Sai Narajala, ``Personal AI Agents like OpenClaw Are a Security Nightmare,'' \url{https://blogs.cisco.com/ai/personal-ai-agents-like-openclaw-are-a-security-nightmare} (accessed July 2026).} Rankings and download counts should not substitute for code review, signing, sandboxing, and revocable permissions.

Finally, containment requires separate credentials and memory-write permissions, since a compromised component that can invoke every tool or rewrite shared memory can affect the entire system. Derived summaries should retain source provenance so that claims can be traced after compression.

\paragraph{Execution transparency and auditability.}
User control also requires an inspectable record of system activity. A personal agent should record which components accessed personal information, invoked tools, delegated tasks, changed memory, or exercised permissions, together with the authorization and outcome of each action. Execution provenance connects these records across evidence, tool use, memory, and agent actions, supporting auditing, failure diagnosis, and recovery \citep{wang2026agenttraces}. Users should be able to review these traces, investigate failures, and confirm that revoked channels or permissions are no longer used. Because traces can reveal sensitive information, their access, storage, and retention controls should reflect the sensitivity of the activity they record.

\paragraph{Goals, feedback, and dependence.}
Goal attainment leaves open which goal should govern an action. Immediate requests can conflict with longer-term commitments, and actions that benefit the user can impose costs on others. A personal agent should be designed to surface these conflicts, seek the user's direction on consequential trade-offs, and respect the consent of affected people.

Once a governing goal is selected, the system must still infer from available feedback whether its actions advance that goal. Immediate approval can be an unreliable proxy, as optimizing it may favor flattery, avoidance of hard truths, or increased dependence. Deployed assistants already show sycophancy, and human preference data can reward it \citep{sharma2024towards}. Evaluation should therefore track delayed outcomes and dependence alongside immediate ratings, while inspection, correction, and revocation give users ways to respond when approval and benefit diverge. Preserving agency also requires attention to how delegation affects the user's ability to revise goals, retain capabilities, and choose which parts of an activity to perform.

\paragraph{Multi-agent composition.}
Personal AI becomes a multi-agent system when agents delegate to one another, share memory or services, or interact with agents serving other users. Work on multi-agent safety distinguishes properties visible in isolation from interactive properties, including an agent's capacity to influence others, susceptibility to exploitation, modeling of other agents, and behavior around shared norms \citep{tilli2026agentproperties}. These properties depend on the counterparties and conditions in which agents interact, so evaluation must include varied agents and realistic threat models.

Coordination among agents with complementary skills may also produce system-level capabilities and risks that single-agent evaluation misses. The distributional AGI safety framework addresses this possibility through sandboxed agent economies with governed transactions, reputation management, and oversight \citep{tomasev2025distributional}. Interacting personal models may also support inferences about communities, institutions, and people who never contributed data. Local control over each observation channel leaves these cross-user effects unresolved. Evaluation should therefore test components and their interactions under realistic configurations of delegation, permissions, shared memory, and external tool access.

\paragraph{Revocation, portability, and storage.}
Control must remain available after a channel is enabled. Users need to inspect stored memory, correct inferences, and revoke observation channels \citep{langheinrich2001privacy}; revocation should propagate to dependent summaries, credentials, queued actions, backups, and synchronized copies. Users also need portable copies of data gathered by their devices so they can change software as needed without losing their history. Restrictions in health and wearable ecosystems limit this control, and discontinued hardware can strand data in vendor-controlled channels.\footnote{Recent examples include the Limitless Pendant, which ceased sale when Meta acquired the company in Dec. 2025 (\url{https://techcrunch.com/2025/12/05/meta-acquires-ai-device-startup-limitless/}, accessed July 2026), and Humane's AI Pin, whose servers shut down days after HP's acquisition in Feb. 2025 (\url{https://techcrunch.com/2025/02/18/humanes-ai-pin-is-dead-as-hp-buys-startups-assets-for-116m/}, accessed July 2026).} Open APIs, documented formats, and user-controlled stores reduce this risk, while local-first and privacy-aware architectures keep ordinary use less dependent on a provider \citep{langheinrich2001privacy,kleppmann2019localfirst}. Keeping memory local concentrates security risk on the user's devices, although remote storage can reduce dependence on any one device and support durable backups and synchronization, with every copy subject to equivalent access and deletion protections.

Remote retrieval introduces another privacy problem because access patterns can reveal private information even when records are protected. Oblivious RAM (ORAM) hides memory-access patterns \citep{ostrovsky1990efficient}; Opal applies this approach to personal AI memory so that the storage provider cannot learn which records a query retrieves \citep{kaviani2026opal}. These systems illustrate how storage architecture shapes privacy when personal memory extends beyond the user's device.

\section{Limitations}
Several limitations follow from the framework's current stage.

Cooperative observation depends on consistent user feedback, which can be burdensome to provide. Privacy concerns can also limit what users share, and outcomes may depend on several active observation channels. A failure may therefore provide little guidance about which channel needs correction. Sparse or inconsistent feedback also makes channel changes harder to interpret. Future studies can combine channel-level provenance with controlled ablations to isolate these effects.

Reflective benefit $U_t$, trust, and task relevance require operational definitions specific to the user, task, and evaluation horizon. The framework also identifies task-conditioned context allocation as a design problem. Comparing allocation methods under processing and disclosure constraints is a separate empirical agenda.

The delegated task distribution may change as users learn which tasks the system handles well. This feedback between system behavior and later inputs is related to performative prediction \citep{perdomo2020performative} and complicates performance comparisons over time.

The behavioral effect of Organizm's coupled agent and memory hierarchies remains unevaluated relative to flat memory or a single agent operating over the same files. Future ablations should test whether the coupling helps and by how much.

Finally, the six-month deployment sketch covers one user under naturalistic conditions. The observed reporting frequency, channel additions, and corrections do not identify causal effects or support generalization, which require controlled multi-user studies.

\section{Conclusion}

Personal agents depend on context relevant to the user's goals, yet much of that context remains outside the system even as devices capture richer traces of daily life. As observation grows richer, task-conditioned context allocation determines which sources and representations enter a decision under processing and disclosure constraints. Cooperative observation describes the resulting feedback loop: the system acts on what it can observe, the user evaluates the result, and that evaluation shapes what the system may observe next. Whether the user maintains a channel depends on the system's usefulness and trustworthiness, the channel's effort and privacy costs, and the control the user retains.

Evaluation should examine whether task-relevant context improves assistance, how assistance quality affects later sharing, correction, and revocation, and how the system's knowledge and the user's self-knowledge change over time.

Organizm implements the loop through manual reports, user-owned memory, and iterative planning. Its six-month deployment records changes in reporting and planning inputs. The single-user record cannot determine why these changes occurred; controlled multi-user studies are needed to test these causes. As sensing becomes more continuous, open software, local-first memory, portable data, and auditable execution can support user control. Future agents should apply personal knowledge to user-chosen goals within user-set delegation boundaries. Users should see and control what personal AI observes, infers, remembers, delegates, and does.

\begin{acks}
This research was supported by the Alberta Machine Intelligence Institute (Amii) and the Canada CIFAR AI Chairs Program. We also thank colleagues at the Network for Applied Technology (NAT), including Yash Mouje, Liliya Eghdamian, Qendrim Beka, and Eric Fung, for their comments and support.
\end{acks}

\bibliographystyle{ACM-Reference-Format}
\bibliography{references}

@article{stray2020aligning,
  title = {Aligning {AI} Optimization to Community Well-Being},
  author = {Stray, Jonathan},
  journal = {International Journal of Community Well-Being},
  volume = {3},
  number = {4},
  pages = {443--463},
  year = {2020},
  doi = {10.1007/s42413-020-00086-3},
  url = {https://doi.org/10.1007/s42413-020-00086-3}
}

@article{boerman2017online,
  title = {Online Behavioral Advertising: A Literature Review and Research Agenda},
  author = {Boerman, Sophie C. and Kruikemeier, Sanne and Zuiderveen Borgesius, Frederik J.},
  journal = {Journal of Advertising},
  volume = {46},
  number = {3},
  pages = {363--376},
  year = {2017},
  doi = {10.1080/00913367.2017.1339368},
  url = {https://doi.org/10.1080/00913367.2017.1339368}
}

@misc{stray2021optimizingforaligningrecommender,
  title = {What are you optimizing for? Aligning Recommender Systems with Human Values},
  author = {Stray, Jonathan and Vendrov, Ivan and Nixon, Jeremy and Adler, Steven and Hadfield-Menell, Dylan},
  year = {2021},
  url = {https://arxiv.org/abs/2107.10939}
}

@article{stray2024building,
  title = {Building Human Values into Recommender Systems: An Interdisciplinary Synthesis},
  author = {Stray, Jonathan and Halevy, Alon and Assar, Parisa and Hadfield-Menell, Dylan and Boutilier, Craig and Ashar, Amar and Bakalar, Chloe and Beattie, Lex and Ekstrand, Michael and Leibowicz, Claire and Moon Sehat, Connie and Johansen, Sara and Kerlin, Lianne and Vickrey, David and Singh, Spandana and Vrijenhoek, Sanne and Zhang, Amy and Andrus, McKane and Helberger, Natali and Proutskova, Polina and Mitra, Tanushree and Vasan, Nina},
  journal = {ACM Transactions on Recommender Systems},
  volume = {2},
  number = {3},
  pages = {1--57},
  year = {2024},
  doi = {10.1145/3632297},
  url = {https://doi.org/10.1145/3632297}
}

@article{kaelbling1998planning,
  title = {Planning and Acting in Partially Observable Stochastic Domains},
  author = {Kaelbling, Leslie Pack and Littman, Michael L. and Cassandra, Anthony R.},
  journal = {Artificial Intelligence},
  volume = {101},
  number = {1--2},
  pages = {99--134},
  year = {1998},
  doi = {10.1016/S0004-3702(98)00023-X},
  url = {https://doi.org/10.1016/S0004-3702(98)00023-X}
}

@inproceedings{hadfieldmenell2016cirl,
  title = {Cooperative Inverse Reinforcement Learning},
  author = {Hadfield-Menell, Dylan and Dragan, Anca and Abbeel, Pieter and Russell, Stuart},
  booktitle = {Advances in Neural Information Processing Systems 29},
  year = {2016},
  url = {https://proceedings.neurips.cc/paper/2016/hash/c3395dd46c34fa7fd8d729d8cf88b7a8-Abstract.html}
}

@inproceedings{christiano2017deep,
  title = {Deep Reinforcement Learning from Human Preferences},
  author = {Christiano, Paul F. and Leike, Jan and Brown, Tom B. and Martic, Miljan and Legg, Shane and Amodei, Dario},
  booktitle = {Advances in Neural Information Processing Systems 30},
  year = {2017},
  url = {https://proceedings.neurips.cc/paper/2017/hash/d5e2c0adad503c91f91df240d0cd4e49-Abstract.html}
}

@inproceedings{ouyang2022training,
  title = {Training Language Models to Follow Instructions with Human Feedback},
  author = {Ouyang, Long and Wu, Jeff and Jiang, Xu and Almeida, Diogo and Wainwright, Carroll L. and Mishkin, Pamela and Zhang, Chong and Agarwal, Sandhini and Slama, Katarina and Ray, Alex and Schulman, John and Hilton, Jacob and Kelton, Fraser and Miller, Luke and Simens, Maddie and Askell, Amanda and Welinder, Peter and Christiano, Paul and Leike, Jan and Lowe, Ryan},
  booktitle = {Advances in Neural Information Processing Systems 35},
  pages = {27730--27744},
  year = {2022},
  doi = {10.52202/068431-2011},
  url = {https://doi.org/10.52202/068431-2011}
}

@inproceedings{li2010stage,
  title = {A Stage-Based Model of Personal Informatics Systems},
  author = {Li, Ian and Dey, Anind K. and Forlizzi, Jodi},
  booktitle = {Proceedings of the SIGCHI Conference on Human Factors in Computing Systems},
  pages = {557--566},
  year = {2010},
  doi = {10.1145/1753326.1753409},
  url = {https://doi.org/10.1145/1753326.1753409}
}

@article{chopra2025engagements,
  title = {Engagements with Generative {AI} and Personal Health Informatics: Opportunities for Planning, Tracking, Reflecting, and Acting around Personal Health Data},
  author = {Chopra, Shaan and Juarez, Katherine and Fogarty, James and Munson, Sean A.},
  journal = {Proceedings of the ACM on Interactive, Mobile, Wearable and Ubiquitous Technologies},
  volume = {9},
  number = {3},
  article = {75},
  pages = {1--33},
  year = {2025},
  doi = {10.1145/3749503},
  url = {https://doi.org/10.1145/3749503}
}

@article{harkin2016monitoring,
  title = {Does Monitoring Goal Progress Promote Goal Attainment? A Meta-Analysis of the Experimental Evidence},
  author = {Harkin, Benjamin and Webb, Thomas L. and Chang, Betty P. I. and Prestwich, Andrew and Conner, Mark and Kellar, Ian and Benn, Yael and Sheeran, Paschal},
  journal = {Psychological Bulletin},
  volume = {142},
  number = {2},
  pages = {198--229},
  year = {2016},
  doi = {10.1037/bul0000025},
  url = {https://doi.org/10.1037/bul0000025}
}

@article{bolger2003diary,
  title = {Diary Methods: Capturing Life as It Is Lived},
  author = {Bolger, Niall and Davis, Angelina and Rafaeli, Eshkol},
  journal = {Annual Review of Psychology},
  volume = {54},
  pages = {579--616},
  year = {2003},
  doi = {10.1146/annurev.psych.54.101601.145030},
  url = {https://doi.org/10.1146/annurev.psych.54.101601.145030}
}

@inproceedings{javed2024bigworld,
  title = {The Big World Hypothesis and its Ramifications for Artificial Intelligence},
  author = {Javed, Khurram and Sutton, Richard S.},
  booktitle = {Finding the Frame: An RLC Workshop},
  year = {2024},
  url = {https://openreview.net/forum?id=Sv7DazuCn8}
}

@inproceedings{joren2025sufficient,
  title = {Sufficient Context: A New Lens on Retrieval-Augmented Generation Systems},
  author = {Joren, Hailey and Zhang, Jianyi and Ferng, Chun-Sung and Juan, Da-Cheng and Taly, Ankur and Rashtchian, Cyrus},
  booktitle = {International Conference on Learning Representations},
  year = {2025},
  url = {https://openreview.net/forum?id=Jjr2Odj8DJ}
}

@inproceedings{langheinrich2001privacy,
  title = {Privacy by Design: Principles of Privacy-Aware Ubiquitous Systems},
  author = {Langheinrich, Marc},
  booktitle = {UbiComp 2001: Ubiquitous Computing},
  series = {Lecture Notes in Computer Science},
  volume = {2201},
  pages = {273--291},
  year = {2001},
  publisher = {Springer},
  doi = {10.1007/3-540-45427-6_23},
  url = {https://doi.org/10.1007/3-540-45427-6_23}
}

@article{parisi2019continual,
  title = {Continual Lifelong Learning with Neural Networks: A Review},
  author = {Parisi, German I. and Kemker, Ronald and Part, Jose L. and Kanan, Christopher and Wermter, Stefan},
  journal = {Neural Networks},
  volume = {113},
  pages = {54--71},
  year = {2019},
  doi = {10.1016/j.neunet.2019.01.012},
  url = {https://doi.org/10.1016/j.neunet.2019.01.012}
}

@inproceedings{kleppmann2019localfirst,
  title = {Local-First Software: You Own Your Data, in Spite of the Cloud},
  author = {Kleppmann, Martin and Wiggins, Adam and van Hardenberg, Peter and McGranaghan, Mark},
  booktitle = {Proceedings of the 2019 ACM SIGPLAN International Symposium on New Ideas, New Paradigms, and Reflections on Programming and Software},
  pages = {154--178},
  year = {2019},
  doi = {10.1145/3359591.3359737},
  url = {https://doi.org/10.1145/3359591.3359737}
}

@article{gemmell2006mylifebits,
  title = {MyLifeBits},
  author = {Gemmell, Jim and Bell, Gordon and Lueder, Roger},
  journal = {Communications of the ACM},
  volume = {49},
  number = {1},
  pages = {88--95},
  year = {2006},
  doi = {10.1145/1107458.1107460},
  url = {https://doi.org/10.1145/1107458.1107460}
}

@inproceedings{hodges2006sensecam,
  title = {SenseCam: A Retrospective Memory Aid},
  author = {Hodges, Steve and Williams, Lyndsay and Berry, Emma and Izadi, Shahram and Srinivasan, James and Butler, Alex and Smyth, Gavin and Kapur, Narinder and Wood, Ken},
  booktitle = {UbiComp 2006: Ubiquitous Computing},
  series = {Lecture Notes in Computer Science},
  volume = {4206},
  pages = {177--193},
  year = {2006},
  publisher = {Springer},
  doi = {10.1007/11853565_11},
  url = {https://doi.org/10.1007/11853565_11}
}

@inproceedings{park2023generative,
  title = {Generative Agents: Interactive Simulacra of Human Behavior},
  author = {Park, Joon Sung and O'Brien, Joseph and Cai, Carrie Jun and Morris, Meredith Ringel and Liang, Percy and Bernstein, Michael S.},
  booktitle = {Proceedings of the 36th Annual ACM Symposium on User Interface Software and Technology},
  pages = {1--22},
  year = {2023},
  doi = {10.1145/3586183.3606763},
  url = {https://doi.org/10.1145/3586183.3606763}
}

@inproceedings{xu2025goals,
  title = {From Goals to Actions: Designing Context-Aware {LLM} Chatbots for New Year's Resolutions},
  author = {Xu, Yan and Jones, Brennan and Nguyen, Hannah and Li, Qisheng and Scherer, Stefan},
  booktitle = {Proceedings of the 7th ACM Conference on Conversational User Interfaces},
  series = {CUI '25},
  articleno = {56},
  numpages = {17},
  year = {2025},
  publisher = {Association for Computing Machinery},
  address = {New York, NY, USA},
  doi = {10.1145/3719160.3736637},
  url = {https://doi.org/10.1145/3719160.3736637}
}

@article{clark1998extended,
  title = {The Extended Mind},
  author = {Clark, Andy and Chalmers, David J.},
  journal = {Analysis},
  volume = {58},
  number = {1},
  pages = {7--19},
  year = {1998},
  doi = {10.1093/analys/58.1.7},
  url = {https://doi.org/10.1093/analys/58.1.7}
}

@article{conant1970good,
  title = {Every Good Regulator of a System Must Be a Model of That System},
  author = {Conant, Roger C. and Ashby, W. Ross},
  journal = {International Journal of Systems Science},
  volume = {1},
  number = {2},
  pages = {89--97},
  year = {1970},
  doi = {10.1080/00207727008920220},
  url = {https://doi.org/10.1080/00207727008920220}
}

@book{minsky1986society,
  title = {The Society of Mind},
  author = {Minsky, Marvin},
  year = {1986},
  publisher = {Simon \& Schuster},
  address = {New York},
  isbn = {978-0-671-60740-1}
}

@incollection{goodhart1984monetary,
  title = {Problems of Monetary Management: The {UK} Experience},
  author = {Goodhart, C. A. E.},
  booktitle = {Monetary Theory and Practice},
  pages = {91--121},
  year = {1984},
  publisher = {Macmillan Education UK},
  address = {London},
  doi = {10.1007/978-1-349-17295-5_4},
  url = {https://doi.org/10.1007/978-1-349-17295-5_4}
}

@book{cover2006elements,
  title = {Elements of Information Theory},
  author = {Cover, Thomas M. and Thomas, Joy A.},
  year = {2006},
  publisher = {Wiley-Interscience},
  edition = {2nd},
  address = {Hoboken, NJ},
  doi = {10.1002/047174882X}
}

@misc{zhao2026clawtrap,
  title = {{ClawTrap}: A {MITM}-Based Red-Teaming Framework for Real-World {OpenClaw} Security Evaluation},
  author = {Zhao, Haochen and Cui, Shaoyang},
  year = {2026},
  url = {https://arxiv.org/abs/2603.18762}
}

@book{polanyi1966tacit,
  title = {The Tacit Dimension},
  author = {Polanyi, Michael},
  year = {1966},
  publisher = {Doubleday},
  address = {Garden City, NY},
  note = {Terry Lectures, Yale University}
}

@misc{pinai2025god,
  title = {{GOD} model: Privacy Preserved {AI} School for Personal Assistant},
  author = {{PIN AI Team}},
  year = {2025},
  url = {https://arxiv.org/abs/2502.18527}
}

@article{genewein2015bounded,
  title = {Bounded Rationality, Abstraction, and Hierarchical Decision-Making: An Information-Theoretic Optimality Principle},
  author = {Genewein, Tim and Leibfried, Felix and Grau-Moya, Jordi and Braun, Daniel Alexander},
  journal = {Frontiers in Robotics and AI},
  volume = {2},
  article = {27},
  year = {2015},
  doi = {10.3389/frobt.2015.00027},
  url = {https://doi.org/10.3389/frobt.2015.00027}
}

@inproceedings{tishby1999information,
  title = {The Information Bottleneck Method},
  author = {Tishby, Naftali and Pereira, Fernando C. and Bialek, William},
  booktitle = {Proceedings of the 37th Annual Allerton Conference on Communication, Control, and Computing},
  pages = {368--377},
  year = {1999},
  url = {https://arxiv.org/abs/physics/0004057}
}

@inproceedings{sharma2024towards,
  title = {Towards Understanding Sycophancy in Language Models},
  author = {Sharma, Mrinank and Tong, Meg and Korbak, Tomasz and Duvenaud, David and Askell, Amanda and Bowman, Samuel R. and Cheng, Newton and Durmus, Esin and Hatfield-Dodds, Zac and Johnston, Scott R. and Kravec, Shauna and Maxwell, Timothy and McCandlish, Sam and Ndousse, Kamal and Rausch, Oliver and Schiefer, Nicholas and Yan, Da and Zhang, Miranda and Perez, Ethan},
  booktitle = {The Twelfth International Conference on Learning Representations},
  year = {2024},
  url = {https://arxiv.org/abs/2310.13548}
}

@article{willett2023highperformance,
  title = {A High-Performance Speech Neuroprosthesis},
  author = {Willett, Francis R. and Kunz, Erin M. and Fan, Chaofei and Avansino, Donald T. and Wilson, Guy H. and Choi, Eun Young and Kamdar, Foram and Glasser, Matthew F. and Hochberg, Leigh R. and Druckmann, Shaul and Shenoy, Krishna V. and Henderson, Jaimie M.},
  journal = {Nature},
  volume = {620},
  number = {7976},
  pages = {1031--1036},
  year = {2023},
  doi = {10.1038/s41586-023-06377-x},
  url = {https://doi.org/10.1038/s41586-023-06377-x}
}

@article{wandelt2024internal,
  title = {Representation of Internal Speech by Single Neurons in Human Supramarginal Gyrus},
  author = {Wandelt, Sarah K. and Bj{\aa}nes, David A. and Pejsa, Kelsie and Lee, Brian and Liu, Charles and Andersen, Richard A.},
  journal = {Nature Human Behaviour},
  volume = {8},
  pages = {1136--1149},
  year = {2024},
  doi = {10.1038/s41562-024-01867-y},
  url = {https://doi.org/10.1038/s41562-024-01867-y}
}

@article{tang2023semantic,
  title = {Semantic Reconstruction of Continuous Language from Non-Invasive Brain Recordings},
  author = {Tang, Jerry and LeBel, Amanda and Jain, Shailee and Huth, Alexander G.},
  journal = {Nature Neuroscience},
  volume = {26},
  pages = {858--866},
  year = {2023},
  doi = {10.1038/s41593-023-01304-9},
  url = {https://doi.org/10.1038/s41593-023-01304-9}
}

@inproceedings{perdomo2020performative,
  title = {Performative Prediction},
  author = {Perdomo, Juan and Zrnic, Tijana and Mendler-D{\"u}nner, Celestine and Hardt, Moritz},
  booktitle = {Proceedings of the 37th International Conference on Machine Learning},
  series = {Proceedings of Machine Learning Research},
  volume = {119},
  pages = {7599--7609},
  publisher = {PMLR},
  year = {2020},
  url = {https://proceedings.mlr.press/v119/perdomo20a.html}
}

@misc{kaviani2026opal,
  title = {Opal: Private Memory for Personal {AI}},
  author = {Kaviani, Darya and Ozdarendeli, Alp Eren and Zhu, Jinhao and Ding, Yu and Popa, Raluca Ada},
  year = {2026},
  url = {https://arxiv.org/abs/2604.02522}
}

@misc{wang2026agenttraces,
  title = {From Agent Traces to Trust: A Survey of Evidence Tracing and Execution Provenance in {LLM} Agents},
  author = {Wang, Yiqi and Zhang, Jiaqi and Cai, Taotao and Liu, Zirui and Sun, Qingqiang and Sun, Zequn and Wu, Zhangkai and Dong, Manqing and Zheng, Mingkai and Yin, Xuefei and Zhu, Yanming},
  year = {2026},
  url = {https://arxiv.org/abs/2606.04990}
}

@inproceedings{ostrovsky1990efficient,
  title = {Efficient Computation on Oblivious {RAM}s},
  author = {Ostrovsky, Rafail},
  booktitle = {Proceedings of the Twenty-Second Annual {ACM} Symposium on Theory of Computing},
  pages = {514--523},
  publisher = {ACM},
  year = {1990},
  doi = {10.1145/100216.100289},
  url = {https://doi.org/10.1145/100216.100289}
}

@inproceedings{tilli2026agentproperties,
  title = {Agent Properties for Multi-Agent Safety},
  author = {Tilli, Cecilia Elena},
  booktitle = {ICLR 2026 Workshop on Agents in the Wild},
  year = {2026},
  url = {https://openreview.net/forum?id=00KtDBoKnO}
}

@misc{kim2026haste,
  title = {Why Solve It Twice? Hierarchical Accumulation of Skills for Transfer-Efficient {ML} Engineering},
  author = {Kim, Yongbin and Talebirad, Yashar and Zaiane, Osmar R.},
  year = {2026},
  url = {https://arxiv.org/abs/2606.30911v2},
  note = {Version 2}
}

@misc{tomasev2025distributional,
  title = {Distributional {AGI} Safety},
  author = {Toma{\v{s}}ev, Nenad and Franklin, Matija and Jacobs, Julian and Krier, S{\'e}bastien and Osindero, Simon},
  year = {2025},
  url = {https://arxiv.org/abs/2512.16856v2},
  note = {Version 2, revised May 19, 2026}
}

@misc{talebirad2026hierarchical,
  title = {Toward a Theory of Hierarchical Memory for Language Agents},
  author = {Talebirad, Yashar and Parsaee, Ali and Szepesv\'{a}ri, Csongor Y. and Nadiri, Amirhossein and Za\"{i}ane, Osmar R.},
  year = {2026},
  note = {ICLR 2026 Workshop on Memory for LLM-Based Agentic Systems},
  url = {https://arxiv.org/abs/2603.21564}
}

@article{xiao2025densing,
  title = {Densing Law of {LLM}s},
  author = {Xiao, Chaojun and Cai, Jie and Zhao, Weilin and Lin, Biyuan and Zeng, Guoyang and Zhou, Jie and Zheng, Zhi and Han, Xu and Liu, Zhiyuan and Sun, Maosong},
  journal = {Nature Machine Intelligence},
  volume = {7},
  pages = {1823--1833},
  year = {2025},
  doi = {10.1038/s42256-025-01137-0},
  url = {https://doi.org/10.1038/s42256-025-01137-0}
}

@inproceedings{bian2026scaling,
  title = {Scaling Laws Meet Model Architecture: Toward Inference-Efficient {LLM}s},
  author = {Bian, Song and Yu, Tao and Venkataraman, Shivaram and Park, Youngsuk},
  booktitle = {International Conference on Learning Representations},
  year = {2026},
  url = {https://arxiv.org/abs/2510.18245}
}

\end{document}